\documentclass[runningheads]{llncs}
\usepackage[dvipsnames]{xcolor}
\usepackage{graphicx}

\usepackage{tikz}
\usepackage{comment}
\usepackage{amsmath,amssymb} 
\usepackage{color}
\usepackage{booktabs} 
\usepackage{array}

\let\llncssubparagraph\subparagraph
\let\subparagraph\paragraph
\usepackage[compact]{titlesec}
\let\subparagraph\llncssubparagraph
\newlength{\Oldarrayrulewidth}

\newcommand{\smallsec}[1]{\vspace{0.04in} \noindent {\bf #1.}}
\newcommand{\olga}[1]{{\color{magenta} Olga: #1}}

\DeclareMathOperator*{\argmin}{arg\,min}

\begin{document}
\pagestyle{headings}
\mainmatter
\def\ECCVSubNumber{3301}  

\title{ELUDE: Generating interpretable \\ explanations via a decomposition into \\labelled and unlabelled features}

\titlerunning{ELUDE: Generating interpretable explanations}
%
\author{Vikram V. Ramaswamy \orcidID{0000-0002-0552-5338} \and \\
Sunnie S. Y. Kim\orcidID{0000-0002-8901-7233} \and
Nicole Meister\orcidID{0000-0002-7154-6882} \and \\
Ruth Fong\orcidID{0000-0001-8831-6402} \and
Olga Russakovsky\orcidID{0000-0001-5272-3241}}
\authorrunning{V. V. Ramaswamy et al.}
%
\institute{Princeton University, Princeton NJ 08544, USA\\
\email{vr23@cs.princeton.edu}}
\maketitle

\begin{abstract}
Deep learning models have achieved remarkable success in different areas of machine learning over the past decade; however, the size and complexity of these models make them difficult to understand. 
In an effort to make them more interpretable, several recent works focus on explaining parts of a deep neural network through human-interpretable, semantic attributes. However, it may be impossible to completely explain complex models using only semantic attributes.  In this work, we propose to augment these attributes with a small set of uninterpretable features.   Specifically, we develop a novel explanation framework ELUDE (Explanation via Labelled and Unlabelled DEcomposition) that decomposes a model's prediction into two parts: one that is explainable through a linear combination of the semantic attributes, and another that is dependent on the set of uninterpretable features.  By identifying the latter, we are able to analyze the ``unexplained'' portion of the model, obtaining insights into the information used by the model. We show that the set of unlabelled features can generalize to multiple models trained with the same feature space and compare our work to two popular attribute-oriented methods, Interpretable Basis Decomposition and Concept Bottleneck, and discuss the additional insights ELUDE provides.  
\keywords{interpretability, explainable AI, global explanations}
\end{abstract}

\section{Introduction}
\label{sec:intro}

In the past decade, deep learning has transformed the field of machine learning with its strong predictive capabilities.
As a result of its many recent successes, deep neural networks are increasingly being applied to high-impact, high-risk domains ranging from precision medicine to autonomous driving.
However, the highly parameterized nature of deep learning that fuels its predictive power also makes these networks challenging to understand.
The ability to thoroughly understand these models is crucial for engendering an appropriate level of trust in them as well as for debugging and fixing undesired model behavior.

Recently, a number of works have focused on explaining aspects of deep neural networks, either through highlighting regions in an image that are important for the model's prediction~\cite{zeiler2014visualizing,simonyan2013saliency,zhou2016cam,zhang2016excitation,selvaraju2017gradcam,chattopadhay2018gradcamplusplus,Petsiuk2018rise,fong19understanding}, labelling neurons or combinations of neurons with semantic attributes~\cite{bau2017netdissect,fong2018net2vec,kim2018tcav,zhou2018ibd} or designing model's that are interpretable by design~\cite{chen2018protopnet,nauta2021prototree,koh2020conceptbottleneck}. However, each of these approaches have specific shortcomings: heatmap based explanations only provide local explanations, labelling parts of the network does not allow us to immediately understand how these parts contribute to the final prediction (if at all), and interpretable-by-design methods tend to encode the presence of attributes as a continuous value rather than a binary, discrete one, making it less interpretable. 

\begin{figure}[t!]
\centering
\includegraphics[width=\textwidth]{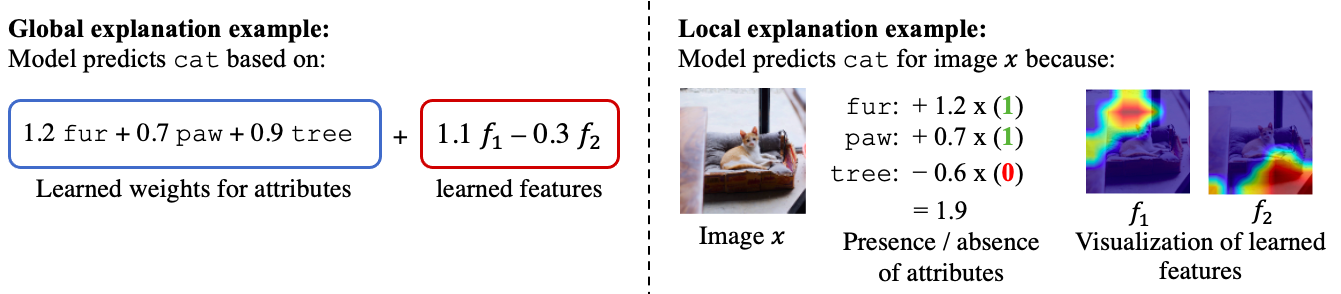}
\caption{Given an image classification model and a set of labelled attributes (like \texttt{fur} or \texttt{paw}), our proposed method ELUDE decomposes the prediction of a class like \texttt{cat} into information gained from labelled attributes and information gained from visual features other than these attributes. This allows us to generate both a global explanation of general attributes important to a class (\emph{left}) as well as a local explanation of how each attribute and learned feature contributes to the overall prediction (\emph{right}).} 
\label{fig:pull_fig}
\end{figure}

In this work, we tackle two challenges that hinder the development of global interpretability methods, and propose our new method: ELUDE (Explanation  via  a  Labelled  and  Unlabelled  DEcomposition). 
The first challenge of global interpretability methods is that they both rely on and are limited by the available labelled attributes.
We propose using the available labelled attributes to the furthest extent possible by first training a interpretable linear classifier on the attributes that simulates the behavior of a blackbox model (i.e. given an input, it produces a similar output along with an explanation using the attributes).
However, these attributes can be insufficient for explaining the entire model, especially as the model grows in complexity (e.g. as the number of classes increases).
Thus, the second challenge we tackle is identifying the remaining portion of the model that is unexplained by the available labelled attributes.
We do so by computing a low-rank subspace within the feature space of the blackbox model, such that, when combined with the labelled attributes, we can fully explain a model's global behavior (see Fig.~\ref{fig:pull_fig}).
Our approach can also be thought of as a \emph{recursive} decomposition, where separating out the semantic information from pre-labelled attributes leaves us with the remaining unexplained components of the network that can then be explained recursively via other methods.

We use ELUDE to explain a scene classification model trained on Places365~\cite{zhou2017places} as well as an interpretable by design Concept Bottleneck model~\cite{koh2020conceptbottleneck} trained on the CUB dataset~\cite{WahCUB_200_2011} and summarize our key takeaways: 
\begin{itemize}
\vspace{-\topsep}
\setlength{\itemsep}{0pt}
\setlength{\parskip}{0pt}
\setlength{\parsep}{0pt}
    \item[1.] The percentage of the model explained with just the attributes reduces as the model increases in complexity. We consider the scene classification model at three levels of granularity (2-way classifier: \texttt{indoor} vs. \texttt{outdoor}, 16-way classifier: scene categories, like \texttt{home-hotel}, \texttt{workplace}, etc, and 365-way classifier: individual scenes, like \texttt{bedroom}, \texttt{kitchen}, etc), and show that the fraction of the model explained using just the attributes reduces as the model increases in complexity. 
    \item[2.] ELUDE can significantly reduce the complexity of the uninterpretable feature space. For the 16-way scene classification model, only a 8-dimensional space is needed to complete the explanation, similarly, for the CUB model, only a 16-dimensional space is needed to complete the explanation.
    \item[3.]  Additional useful concepts can be learned from analysing the low-rank features. By visualizing images that highly activate different dimensions within the low-rank space, we can identify concepts such as ``bowling alleys'' or ``castle-like buildings'' that could be used by the model. 
    \item[4.] The low-rank subspace generalizes. A common low-rank space can be learned across multiple models trained on the same feature space, which can also be used to explain unseen models. 
\end{itemize}
Altogether, these findings show that ELUDE constructs a simple explanation that allows users to explain a blackbox model with known attributes and condenses the remainder of the model into a simple space which can be further explored, taking a key step towards completely explaining a blackbox model. 

\section{Related work}
\label{sec:related}


A popular set of interpretability methods~\cite{zeiler2014visualizing,simonyan2013saliency,zhou2016cam,zhang2016excitation,selvaraju2017gradcam,chattopadhay2018gradcamplusplus,Petsiuk2018rise,fong19understanding} focus on characterizing the image regions responsible for a model's decision via a heatmap visualization.
Another work (LIME~\cite{ribeiro2016lime}) does this by identifying relevant superpixels in an image.
Unlike these local methods, which aim to explain individual predictions, our method is global in that it attempts to explain the entire model, while being able to simultaneously produce local explanations. 

Some global interpretability works~\cite{bau2017netdissect,fong2018net2vec,kim2018tcav,zhou2018ibd} attempt to identify regions within a network that correspond to specific attributes, either through specific neurons (NetDissect~\cite{bau2017netdissect}) or through linear combinations of neurons (Net2Vec~\cite{fong2018net2vec} and concept activation vectors~\cite{kim2018tcav}). 
We also use pre-labelled attributes to give an explanation, but differ in two key ways: we focus on explaining the model's predictions, rather than intermediate representations within the model and we derive a subspace of the feature space that contributes to the prediction while not being correlated with any of the attributes.
Interpretable Basis Decomposition~\cite{zhou2018ibd} is most similar to our work, as it also decomposes a prediction into a linear sum of attributes and a residual; however, we differ from their work in that we allow negative weights, which allow for more of the model to be explained using attributes, and we work at the feature space after pooling as opposed to before.
Moreover, we extract a small dimension subspace of the feature space that contains the visual cues used by the blackbox model other than the attributes. 
 
Recently, several models have been proposed, mostly for fine-grain visual recognition, that learn a model that is interpretable-by-design~\cite{chen2018protopnet,koh2020conceptbottleneck,nauta2021prototree}.
The first class of models (ProtoPNet~\cite{chen2018protopnet}, ProtoTree~\cite{nauta2021prototree}) learns a set of unlabelled prototypes from the training set and computes an input image's similarity to the learned prototypes in order to make predictions.
More similar to our method are the second class of models (Concept Bottleneck~\cite{koh2020conceptbottleneck} and Contextual Semantic Interpretability~\cite{marcos_accv_2020}) which first learn to predict a set of labelled attributes and then learn a second, discriminative model that uses the predicted attributes as input. We differ from these in that we use binary ground truth attribute labels rather than using a model's (fractional) prediction for these attributes. 


Finally, our approach is related in spirit to works on diagnosing and analyzing errors in computer vision tasks such as object detection~\cite{hoiem2012error,bolya2020tide} and action recognition~\cite{sigurdsson2017action}.
We aim to develop a tool that allows researchers to diagnose what parts of the network are unable to be explained using the labelled attributes.



\section{Method}
\label{sec:methods}

\noindent Given a CNN-based image classification model $F$ and a set of semantic attributes $A$, our goal is to create a global interpretability method that allows us to understand the entire model as well as simulate the model on a new input. 
To do so, we assume we have access to a set of images (typically distinct from the training set of $F$), for which we have  $K$ binary semantic attribute labels $A(x) \in \{0,1\}^K$ annotated for each image $x$.
Like previous works~\cite{koh2020conceptbottleneck,fong2018net2vec,zhou2018ibd,kim2018tcav}, we also assume that the labelled attributes are encoded linearly within the feature space (typically, the penultimate layer of $F$); thus, we provide an explanation that is a linear combination of these attributes.



\smallsec{Explaining using attributes} We begin by explaining as much of the model as possible using the semantic attributes. 
Given a set of images, we can train a linear model\footnotemark \ $W_A$ directly on the semantic attributes $A(x)$ to approximate the network predictions $F(x)$. This allow us to both see how well the output distribution of the original network can be approximated with the semantic attributes and evaluate the contribution of each using the linear weights of $W_A$. In line with ensuring that the explanation is as human-interpretable as possible, we train this linear model with an $L1$ sparsity penalty on the coefficients.
\footnotetext{We chose to use a linear model as our explanation to ensure human understandability.}

However, this forces the explanation to rely \emph{exclusively} on the semantic attributes; i.e., it does not allow for the possibility that the network is using any additional information. 
As the model grows in complexity (e.g., as the number of classes increases), using only the given attributes to explain the entire model becomes insufficient.
Instead, we attempt to derive a simple subspace from the feature space $f$ (typically, the penultimate layer of $F$) that, when added to our set of attributes, allows us to more fully explain the model.  


\smallsec{Contribution of the remaining features}
We propose to learn our explanation as a sum of two linear models - one that relies exclusively on $A$ and another that relies exclusively on the uninterpretable features $f$.
When training our explanation, we want our explanation to mimic the blackbox model's output as far as possible; thus, we minimize the cross-entropy loss between the model's predicted class and the output of the explanation.\footnote{We choose to explain the discrete labels outputted by the model rather than the scores output in order to keep the explanation human-understandable.}
More precisely, let $x_1, x_2, \ldots, x_N$ denote the set of images, and $\hat{y}_1, \hat{y}_2, \ldots, \hat{y}_N$ denote the discrete class labels predicted for each image.
Then, we want to learn linear models $W_A, W_f$\footnotemark:
\begin{align}
    \label{eq:false}
    \argmin_{W_A, W_f} & \sum_i \mbox{CrossEntropy}\big(\hat{y}_i, W_A^TA(x_i) + W_f^Tf(x_i)\big)
\end{align}
However, this minimization can be achieved by simply utilizing the linear layer of $F$ as $W_f$, ignoring the attributes.
\footnotetext{To simplify notation, we assume that the bias term is represented within $A(X)$ and $f(x)$, i.e., we assume that the final value within these is 1 for all images $x_i$}
We thus train these in sequence: We first learn $W^T_AA(x)$ to approximate $\hat{y}$ as well as possible, and then learn $W_f$.

\begin{figure}[t]
    \centering
    \includegraphics[width=\linewidth]{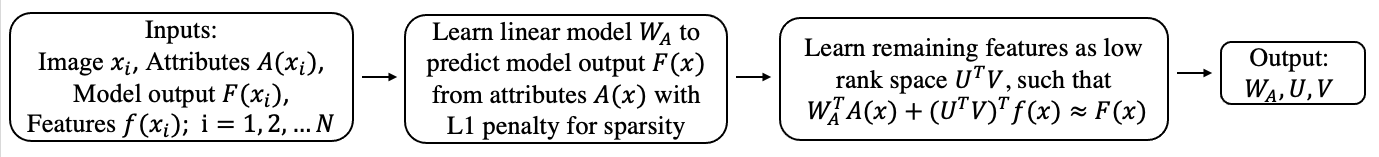}
    \caption{A flowchart of ELUDE: Explanation via a Labelled and Unlabelled DEcomposition.} 
    \label{fig:methods}

\end{figure}

Another concern is that $W_f$ may be as hard to explain as the original black box model.
We address this by limiting the rank of $W_f$, i.e., we force $W_f = U^TV$, with $U$ having a low rank $r$.
Then, our final optimization is as follows:
\begin{align}
\label{eq:objection_single}
\begin{split}
\argmin_{W_A} & \sum_i \mbox{CrossEntropy}\big(\hat{y}_i, W_A^TA(x_i) \big) + \lambda_1 ||W_A||_1 \\
\argmin_{ U, V| rank(U)=r} & \sum_i \mbox{CrossEntropy} \big(\hat{y}_i, W_A^TA(x_i) + (U^TV)^Tf(x_i)\big) 
\end{split}
\end{align}
This process is depicted in Fig.~\ref{fig:methods}. The features within the model that do not correspond to the attributes in $A$ can be represented as $Uf$, which, due to the low rank of $U$, is an easier-to-understand feature space. The final explanation can be considered as a weighted combination of the attributes $A$ (using weights $W_A$) and the new learned features $Uf$ (using weights $V$). We refer to the method to computing $W_A, U$, and $V$ using Eq.~\ref{eq:objection_single} as ELUDE: Explanation via a Labelled and Unlabelled DEcomposition. 

\smallsec{Choosing the rank} 
The quality of the overall explanation depends on the rank of $U$. 
A higher rank allows us to explain more of the model but also yields more directions of $Uf$ that need to be interpreted.
A rank that is too low explains the model less well; furthermore, it may make the directions of $Uf$ less interpretable since each direction is maybe forced to encode multiple concepts. 
To strike the right balance, we first compute the minimum rank at which the uninterpretable features alone can mimic the model, i.e., we find the minimum rank $r_{all}$ that is necessary for a rank $r_{all}$ projection $U$ to explain the model.
Assuming that we now use $r_A$ attributes from $A$ in $W_A$ ($r_A$ might be less than $|A|$, since we train $W_A$ with $L1$ regularization), we choose the rank of $U$ in our final explanation to be roughly $r_{all}-r_A$.\footnote{In practice, we choose the rank to be a power of 2 close to $r_{all}-r_A$.} 
We note that learning $W_A$ with an $L1$ penalty encourages independence between the attributes with non-zero weights, which further gives credence to this choice of rank. 

The combined linear explanation provided by ELUDE is simple but powerful, allowing us to explain both global and local properties of the model. At a global level, ELUDE highlights a set of attributes along with their relative importances for each target class, while for an individual image, we are  able to identify regions of the image not corresponding to the labelled attributes that are important to the model's prediction. 
Additionally, we can quantify the percentage of a model that \emph{can} and \emph{cannot} be explained by the set of labelled attributes.
The characterization of the uninterpretable space as a small rank subspace also allows us to gain insights into other potential concepts being used within the model.
Moreover, as we show in Sec.~\ref{sec:beyond_single}, this learned space can even generalize to multiple linear models trained on the features in $f$. 

\section{Explaining blackbox scene classification models}
\label{sec:places}
We examine the effectiveness of ELUDE in two settings: first, in this section, we explain a standard CNN~\cite{he2016resnet} trained for scene classification on Places365~\cite{zhou2017places} and second, in Sec.~\ref{sec:cub}, we explain both an interpretable-by-design Concept Bottleneck~\cite{koh2020conceptbottleneck} model as well as a standard CNN on the CUB dataset~\cite{WahCUB_200_2011}. 

In this section, we explain a ResNet18~\cite{he2016resnet} model trained on Places365~\cite{zhou2017places}. This model takes as input an image and outputs a 365 dimension vector, with the predicted probabilities of the image belonging to each class. We consider these predictions at different granularities: first, we group all scenes into 2 classes corresponding to \texttt{indoor} and \texttt{outdoor} scenes (Sec.~\ref{sec:places_binary}), second, we group all scenes into one of 16 categories, as proposed by the dataset creators, such as \texttt{home/hotel}, \texttt{indoor-cultural}, \texttt{forest/field/jungle}, etc. (Sec.~\ref{sec:places_16}), and finally, we use all 365 classes (Sec.~\ref{sec:ibd_comp}). This allows us to examine how ELUDE adapts to the increase in complexity of the blackbox model. We also consider a multi-head classifier to understand how the learned low-rank space generalises. 

\smallsec{Semantic attributes}  We use the ADE20k~\cite{zhou2017ade20k,zhou2019ade20k_ijcv} dataset (license: BSD 3-Clause) to evaluate our method, splitting the dataset images randomly into train (60\%, 11839 images), val (20\%, 3947 images), and test (20\%, 3947 images) splits to use for the classifiers,  using the new training set for learning the explanations, validation set for tuning the hyperparameters (e.g., the $L1$ regularization weight, and the rank $r$) and test set for reporting our findings. ADE20k is a subset of the Broden dataset~\cite{bau2017netdissect} which is labelled with a variety of attributes. 
For the attributes $A$, we use all object and part Broden attributes that occur in at least 150 of the ADE20k training images. This comprises of 112 attributes.

\smallsec{Implementation details}
We use the features output after the average pool layer in the ResNet18 model as $f$. We train $W_A$ with scikit-learn's~\cite{scikit-learn} LogisticRegression model with an $L1$ penalty and a liblinear solver.\footnote{We choose the regularization weight by considering the change in accuracy as the parameter changes and choosing the point where the accuracy levels. More details are present in the {appendix}. } When training $W_f = U^TV$, we use an Adam~\cite{KB14Adam} optimizer and a learning rate 1e-4.

\smallsec{Metrics}
We want our explanation to match the model's output as much as possible, while remaining easy to understand. Hence, we report both the number of attributes our model uses along with the number of uninterpretable features (the rank of $U$), as well as the fraction of model's outputs we are able to explain.  


\subsection{Explaining a binary classifier}
\label{sec:places_binary}
In this section, we attempt to explain when a Resnet18 model trained on Places365 predicts \texttt{indoor} versus \texttt{outdoor} scenes. 

\smallsec{Establishing an upper bound} An issue with all interpretability methods that use a different dataset to explain their model rather than the one the model was trained on is that this can reduce the fraction of the model being explained. Here, we use the ADE20k dataset to explain a model trained on Places365. To quantify this limit, we train a linear layer on the feature space $f(x)$ to match the black box model's output $F(x)$ using images $\{x_i\}_{i=1}^N$ from the ADE20k dataset; this establishes an upper bound of $97.7\%$.

\smallsec{Explaining with labelled attributes}
We first learn a sparse linear model with the 112 attributes from ADE20k to match the model output. This explanation has just 6 non-zero coefficients, and is able to explain 95.7\% of the model's outputs, very close to our upper bound of 97.7\%. These attributes are (in decreasing coefficient magnitude\footnote{Table with actual weights is in the appendix}) \texttt{sky}, \texttt{floor}, \texttt{tree},  \texttt{wall}, \texttt{building}, and \texttt{ceiling}. As expected, \texttt{sky}, \texttt{tree}, and \texttt{building} are considered important for \texttt{outdoor} scenes, while the others are considered important for \texttt{indoor} scenes. That is, this simple binary classifier is almost fully explained with just attributes. 

\subsection{Explaining a 16-way classifier}
\label{sec:places_16}
We next use ELUDE to explain a Resnet18 model trained on Places365 to output one of 16 scene categories. This is a slightly more complex model, and we find that the labelled attributes are no longer sufficient to explain the model, requiring us to compute a low-dimensional uninterpretable feature space. 

\smallsec{Establishing an upper bound}
We again begin by computing the upper bound for how well this model can be explained using the ADE20k dataset; and find that the upper bound is 78.0\%. 
While this may seem low, we note that this underscores the difficulty of the task: we are trying to understand a very complex, 16-way deep learning classifier.\footnote{For comparison, note that e.g.,  NetDissect~\cite{bau2017netdissect} and Interpretable Basis Decomposition~\cite{zhou2018ibd} are able to explain only 52.9\% and 65.9\% (numbers taken from \cite{bau2017netdissect,zhou2018ibd}) of the network's neurons respectively}

\begin{table*}[t!]
    \centering
        \caption{We report the true positive rate for the attribute-only explanation ELUDE provides. We see that the explanation is much better for some classes than others, due to the availability of the relevant attributes in our explanation dataset. We also report the five most important attributes for each scene category, based on the \textcolor{NavyBlue}{\textbf{\texttt{positive}}} or \textcolor{red}{\textit{\texttt{negative}}} weight the linear model learned on them (zero-weight attributes always omitted). We see that the attributes correspond well to our intuition: presence of \texttt{building} and \texttt{skyscraper} contributes \textcolor{NavyBlue}{positively} to the score of  \texttt{commercial-buildings/towns}, whereas presence of  \texttt{bed} is contributes \textcolor{red}{negatively} to the score of \texttt{workplace}. }
    \resizebox{0.9\linewidth}{!}{%
    \begin{tabular}{l p{0.5cm} c p{0.5cm} l}
\toprule
Scene group && TPR  && Important attributes \\
\toprule
\texttt{home/hotel} && 99.0 &&  \textcolor{NavyBlue}{\textbf{\texttt{bed}}},  \textcolor{red}{\textit{\texttt{sky}}},  \textcolor{NavyBlue}{\textbf{\texttt{sink}}},  \textcolor{red}{\textit{\texttt{person}}},  \textcolor{red}{\textit{\texttt{tree}}} \\
\texttt{comm-buildings/towns} && 93.5 && \textcolor{NavyBlue}{\textbf{\texttt{building}}},  \textcolor{NavyBlue}{\textbf{\texttt{skyscraper}}},  \textcolor{red}{\textit{\texttt{wall}}},  \textcolor{red}{\textit{\texttt{floor}}},  \textcolor{NavyBlue}{\textbf{\texttt{sidewalk}}} \\
\texttt{water/ice/snow} && 60.6 && \textcolor{red}{\textit{\texttt{wall}}},  \textcolor{red}{\textit{\texttt{building}}},  \textcolor{red}{\textit{\texttt{floor}}},  \textcolor{NavyBlue}{\textbf{\texttt{mountain}}},  \textcolor{red}{\textit{\texttt{road}}} \\
\texttt{forest/field/jungle} && 40.2 && \textcolor{red}{\textit{\texttt{wall}}},  \textcolor{red}{\textit{\texttt{floor}}},  \textcolor{red}{\textit{\texttt{building}}},  \textcolor{NavyBlue}{\textbf{\texttt{tree}}},  \textcolor{red}{\textit{\texttt{road}}} \\
\texttt{workplace} && 14.2 &&  \textcolor{red}{\textit{\texttt{sky}}},  \textcolor{red}{\textit{\texttt{tree}}},  \textcolor{red}{\textit{\texttt{bed}}},  \textcolor{red}{\textit{\texttt{building}}} \\
\texttt{shopping-dining} && 12.4 &&  \textcolor{red}{\textit{\texttt{sky}}},  \textcolor{NavyBlue}{\textbf{\texttt{shelf}}},  \textcolor{NavyBlue}{\textbf{\texttt{person}}},  \textcolor{red}{\textit{\texttt{tree}}},  \textcolor{red}{\textit{\texttt{windowpane}}} \\
\texttt{cultural/historical} && 6.5 && \textcolor{red}{\textit{\texttt{floor}}},  \textcolor{NavyBlue}{\textbf{\texttt{building}}},  \textcolor{red}{\textit{\texttt{car}}},  \textcolor{red}{\textit{\texttt{ceiling}}},  \textcolor{NavyBlue}{\textbf{\texttt{grass}}} \\
\texttt{cabins/gardens/farms} && 4.7 && \textcolor{red}{\textit{\texttt{floor}}},  \textcolor{NavyBlue}{\textbf{\texttt{tree}}},  \textcolor{NavyBlue}{\textbf{\texttt{plant}}},  \textcolor{NavyBlue}{\textbf{\texttt{house}}},  \textcolor{red}{\textit{\texttt{ceiling}}} \\
\texttt{outdoor-transport} && 3.2 &&  \textcolor{red}{\textit{\texttt{floor}}},  \textcolor{red}{\textit{\texttt{wall}}},  \textcolor{NavyBlue}{\textbf{\texttt{car}}},  \textcolor{NavyBlue}{\textbf{\texttt{road}}},  \textcolor{NavyBlue}{\textbf{\texttt{building}}} \\
\texttt{indoor-transport} && 0.0 &&  \textcolor{red}{\textit{\texttt{sky}}},  \textcolor{red}{\textit{\texttt{tree}}},  \textcolor{NavyBlue}{\textbf{\texttt{seat}}} \\
\texttt{indoor-sports/leisure} && 0.0 &&  \textcolor{red}{\textit{\texttt{sky}}} \\
\texttt{indoor-cultural} && 0.0 && \textcolor{red}{\textit{\texttt{sky}}},  \textcolor{red}{\textit{\texttt{tree}}} \\
\texttt{mountains/desert/sky} && 0.0  &&  \textcolor{red}{\textit{\texttt{wall}}},  \textcolor{NavyBlue}{\textbf{\texttt{mountain}}},  \textcolor{red}{\textit{\texttt{building}}} \\
\texttt{outdoor-manmade} && 0.0 && \textcolor{red}{\textit{\texttt{floor}}},  \textcolor{red}{\textit{\texttt{wall}}} \\
\texttt{outdoor-fields/parks} && 0.0 && \textcolor{red}{\textit{\texttt{wall}}},  \textcolor{red}{\textit{\texttt{floor}}} \\
\texttt{industrial-construction} && 0.0 &&  \textcolor{red}{\textit{\texttt{floor}}},  \textcolor{red}{\textit{\texttt{wall}}} \\
\toprule
\end{tabular}
}

    \label{tab:ade20k_impattr}
\end{table*}

\smallsec{Explaining with labelled attributes}
Having established an upper bound, we now learn a sparse linear model $W_A$ to predict the model's output using the labelled attributes. 
Tab.~\ref{tab:ade20k_impattr} shows which attributes are important to each class, along with the true positive rate for each. The explanation matches well with our intuition: for example, for the scene category \texttt{shopping-dining}, the {presence} of attributes \texttt{shelf} and \texttt{person} is important, while the {presence} of \texttt{sky} reduces the score. The true positive rate varies across classes; one potential explanation is that ADE20k is not well balanced, and there are more images from classes like \texttt{home-hotel}. However, even for scene categories with low true positive rates, ELUDE highlights some important attributes: scores for indoor categories, like \texttt{indoor-sports/leisure}, \texttt{indoor-cultural}, \texttt{indoor-transport} all reduce in the presence of \texttt{sky}.  However, on the whole, this linear model only explains 46.2\% of the blackbox model's predictions. Given that our established upper bound is 78.0\%, we can ask what else the model uses in order to make predictions.

\smallsec{Choosing the rank}
As proposed in Sec.~\ref{sec:methods}, we choose the rank for the uninterpretable space by computing the minimum rank required for the feature space to mimic the model's prediction, and subtract the number of attributes in $W_A$. We find that the minimum number of features required ($r_{all}$) is 32. Since the number of attributes with non-zero coefficients in $W_A$ is 29, $r_A = 29$. Hence, the optimal rank is roughly $r_all - r_A = 3$. 

\begin{figure}[t]
\centering
\resizebox{0.9\textwidth}{!}{
\begin{minipage}{.42\textwidth}
  \centering
  \includegraphics[width=\linewidth]{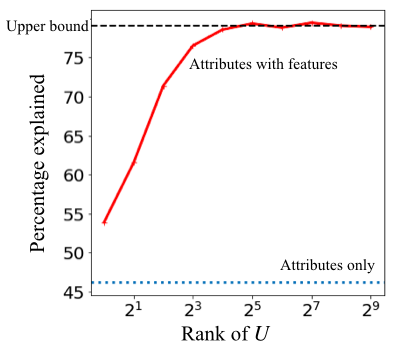}
\end{minipage}%
\begin{minipage}{.58\textwidth}
   \centering
   \includegraphics[width=\linewidth]{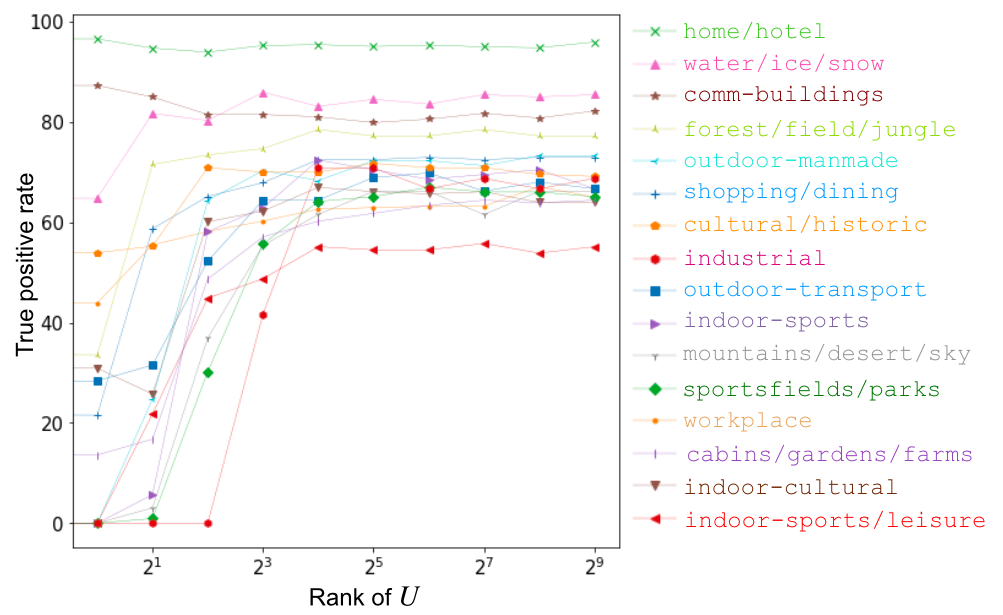}

\end{minipage}}
\caption{For a 16-way scene classification model trained on Places365, we visualize how much of this model is explained through both ADE20k attributes and a low rank feature space. For the overall model \emph{(left)}, we see that with a rank 8 projection, we are able to predict the model's output close to the upper bound. For the individual scene categories (\emph{right}), we see that the true positive rate for each start to stabilize roughly at $r=8$.} 
\label{fig:ade_analysis}
\end{figure}

\begin{figure}[!ht]
    \centering
    \includegraphics[width=0.85\linewidth]{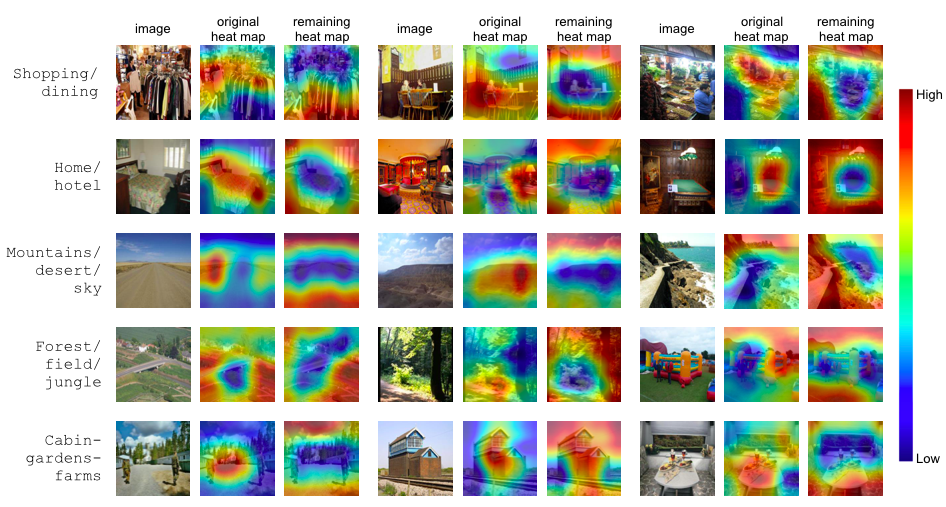}
    \caption{For each scene category, we visualize a set of images, GradCAM on the baseline model (original), as well as on our explanation (remainder) when using ELUDE to explain the 16-way scene classification model. The GradCAM of our explanation shows us how much the model depends on features encoded in these image regions \emph{in addition to} the labelled semantic attributes.
    Here, we display 5 of the 16 categories. In general, we notice that our remaining explanation for the images depends less on the objects at the center of the image, suggesting that our explanation learns to use the salient labelled objects directly from $A$, for example, \texttt{bed} in the first set of images in row 2.}
    \label{fig:cam_ade}
\end{figure}
\begin{figure}[!ht]
    \centering
    \includegraphics[width=0.95\linewidth]{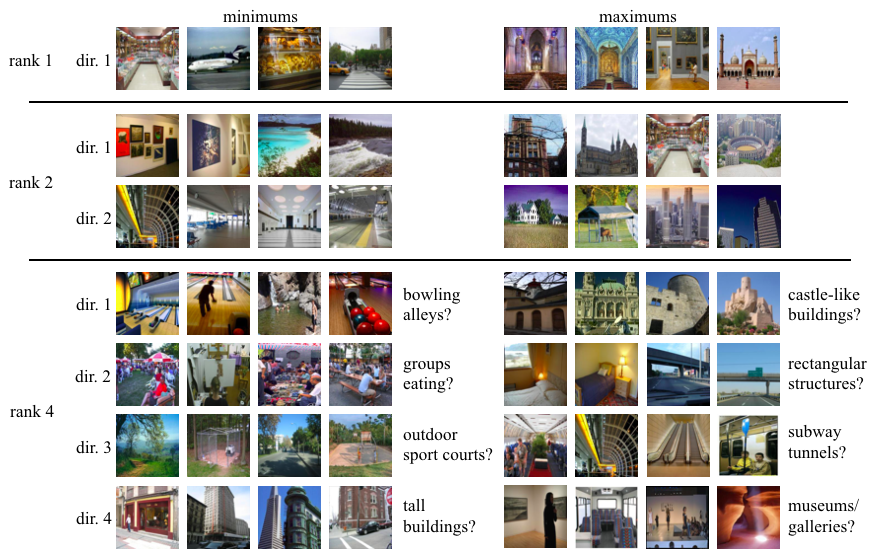}
    \caption{We show images the minimally (left) and maximally (right) activate each learned feature in the low rank subspace for different ranks for the model in Sec.~\ref{sec:places_16}. For lower ranks of 1 and 2, the features still do not correspond to semantic attributes. As the rank increases, directions appear to correspond to certain concepts, e.g., for rank 4, the images that maximally activate direction 1 appear to be castle-like buildings.}
    \label{fig:max_activated}
\end{figure}

\smallsec{Completing the explanation} Following ELUDE, we compute a low-rank projection $U$ from $f$ that allows us to explain more of the model's predictions. Fig~\ref{fig:ade_analysis} (\emph{left}) demonstrates how the fraction of instances we are able to explain changes as we increase the rank of the subspace $U$. With a rank of 4, we are able to explain over 70\% of the model's predictions, and this increases to over 75\% with a rank of 8, which is reasonably close to our upper bound. In Fig~\ref{fig:ade_analysis} (\emph{right}), we view the individual true positive rates for different scene categories and find that  the true positive rates for all the scene categories start to stabilize at $r=8$, which further validates our choice of rank. One reason for the larger rank could be that our set of 29 attributes within $W_A$ do not capture all 16 classes (as seen in Tab.~\ref{tab:ade20k_impattr}), and the unrepresented classes require more information.  

\smallsec{Analysing $Uf$}
In order to analyze the learned projection $U$, we visualize the difference in the GradCAM~\cite{selvaraju2017gradcam} of our explanation $(U^TV)^Tf(x)$ and $F(x)$ (Fig.~\ref{fig:cam_ade}). The GradCAM images for our explanation, which highlight regions within the image that are not described by the attributes in $A$, often focus on the edges of the image, rather than salient objects in the center. This suggests that the model pays attention to some of the context around these annotated objects. 

 We also visualize images that highly activate the space for each dimension, as shown in Fig.~\ref{fig:max_activated}. We see that as the rank of $U$ increases, features in the uninterpretable subspace appear to correspond to semantic concepts; e.g., at  rank 4, images that maximally activate direction 1 consist of monuments, and those that minimally activate direction 2 consist of people eating. Labelling these concepts, and adding them to the list of attributes would allow us to further reduce the rank of the uninterpretable space.

\subsection{Explaining a 365-way classifier.}
\label{sec:ibd_comp}
Finally, we explain the most complex 365-scene classifier with ELUDE and compare the resulting explanation to that of Interpretable Basis Decomposition~\cite{zhou2018ibd}. 
To make a fair comparison, we use the same set of 660 attributes from Broden~\cite{bau2017netdissect} as used in IBD.

We compare the important attributes used for the model's prediction for these two methods in Tab.~\ref{tab:ibd-comp}. We find that the important attributes with positive weights tend to remain the same, however, ELUDE allows us to discover attributes whose presence \textit{reduces} the score for the class. The $L1$ penalty added also allows our explanation to focus more on attributes that occur more, for example, we see that ELUDE does base a lot of predictions on commonly occuring attributes like \texttt{floor} and \texttt{sky}. We describe the computed low-rank uninterpretable subspace for this model in the {appendix}. 

\begin{table}[t]
    \centering
        \caption{We compare the top 6 important attributes (based on the coefficient weight) reported for 6 scenes chosen at random from the 365 Places365 categories. We find that the important attributes tend to agree, but ELUDE also discovers attributes presence significantly reduces the score of a particular class. Moreover, in contrast to the high-dimensional residual provided by IBD, ELUDE provides a low-dimension subspace that summarizes what is left to be explained.}
    \label{tab:ibd-comp}
\resizebox{0.9\linewidth}{!}{%
\begin{tabular}{>{\raggedright\arraybackslash}p{0.18\textwidth}>{\raggedright\arraybackslash}p{0.4\textwidth}>{\raggedright\arraybackslash}p{0.4\textwidth}}
\toprule
scene name & IBD \cite{zhou2018ibd} & ELUDE  \\
\toprule
\begin{tabular}{l}
  \texttt{movie-}\\\texttt{theater}\\\texttt{/indoor}
\end{tabular} & \begin{tabular}{l}
 \textcolor{NavyBlue}{\textbf{\texttt{silver screen}}}, \textcolor{NavyBlue}{\textbf{\texttt{stage}}}, \\ \textcolor{NavyBlue}{\textbf{\texttt{television stand}}}, \textcolor{NavyBlue}{\textbf{\texttt{barrels}}}, \\ \textcolor{NavyBlue}{\textbf{\texttt{tvmonitor}}}, \textcolor{NavyBlue}{\textbf{\texttt{seat}}}
\end{tabular}  & \begin{tabular}{l}
\textcolor{NavyBlue}{\textbf{\texttt{silver screen}}}, \textcolor{NavyBlue}{\textbf{\texttt{microphone}}}, \\  \textcolor{NavyBlue}{\textbf{\texttt{stage}}}, \textcolor{NavyBlue}{\textbf{\texttt{seat}}}, \textcolor{red}{\textit{\texttt{windowpane}}}, \\ \ \textcolor{NavyBlue}{\textbf{\texttt{curtain}}}
\end{tabular}  \\ \hline
\texttt{embassy} & \begin{tabular}{l}
\textcolor{NavyBlue}{\textbf{\texttt{streetlight}}}, \textcolor{NavyBlue}{\textbf{\texttt{windows}}}, \textcolor{NavyBlue}{\textbf{\texttt{balcony}}}, \\  \textcolor{NavyBlue}{\textbf{\texttt{curb}}}, \textcolor{NavyBlue}{\textbf{\texttt{mosque}}}, \textcolor{NavyBlue}{\textbf{\texttt{slats}}} 
\end{tabular}  & \begin{tabular}{l}
 \textcolor{NavyBlue}{\textbf{\texttt{building}}}, \textcolor{NavyBlue}{\textbf{\texttt{stairway}}}, \textcolor{red}{\textit{\texttt{box}}}, \\  \textcolor{NavyBlue}{\textbf{\texttt{board}}}, \textcolor{NavyBlue}{\textbf{\texttt{hedge}}}, \textcolor{red}{\textit{\texttt{floor}}}
\end{tabular}   \\ \hline
\texttt{jail-cell} & \begin{tabular}{l}
\textcolor{NavyBlue}{\textbf{\texttt{cage}}}, \textcolor{NavyBlue}{\textbf{\texttt{toilet}}}, \textcolor{NavyBlue}{\textbf{\texttt{grille door}}}, \\  \textcolor{NavyBlue}{\textbf{\texttt{vent}}}, \textcolor{NavyBlue}{\textbf{\texttt{ticket window}}}, \\ \textcolor{NavyBlue}{\textbf{\texttt{water tank}}}
\end{tabular}   & \begin{tabular}{l}
\textcolor{NavyBlue}{\textbf{\texttt{grille door}}}, \textcolor{NavyBlue}{\textbf{\texttt{bar}}}, \textcolor{red}{\textit{\texttt{painting}}}, \\  \textcolor{NavyBlue}{\textbf{\texttt{sink}}}, \textcolor{NavyBlue}{\textbf{\texttt{bed}}}, \textcolor{red}{\textit{\texttt{sky}}}
\end{tabular}  \\ \hline
\texttt{auditorium} & \begin{tabular}{l}
\textcolor{NavyBlue}{\textbf{\texttt{stage}}}, \textcolor{NavyBlue}{\textbf{\texttt{seat}}}, \textcolor{NavyBlue}{\textbf{\texttt{silver screen}}},  \\  \textcolor{NavyBlue}{\textbf{\texttt{barrels}}}, \textcolor{NavyBlue}{\textbf{\texttt{stalls}}}, \textcolor{NavyBlue}{\textbf{\texttt{grandstand}}}
\end{tabular}  &  \begin{tabular}{l}
\textcolor{NavyBlue}{\textbf{\texttt{seat}}}, \textcolor{NavyBlue}{\textbf{\texttt{stage}}}, \textcolor{red}{\textit{\texttt{painting}}}, \textcolor{NavyBlue}{\textbf{\texttt{piano}}}, \\   \textcolor{NavyBlue}{\textbf{\texttt{spotlight}}}, \textcolor{red}{\textit{\texttt{cabinet}}} 
\end{tabular}  \\ \hline
\begin{tabular}{l}\texttt{science-}\\\texttt{museum}\end{tabular} &  \begin{tabular}{l}
\textcolor{NavyBlue}{\textbf{\texttt{case}}}, \textcolor{NavyBlue}{\textbf{\texttt{wing}}}, \textcolor{NavyBlue}{\textbf{\texttt{drawing}}}, \textcolor{NavyBlue}{\textbf{\texttt{skeleton}}}, \\   \textcolor{NavyBlue}{\textbf{\texttt{video player}}}, \textcolor{NavyBlue}{\textbf{\texttt{bell}}} 
\end{tabular}   &  \begin{tabular}{l}
\textcolor{NavyBlue}{\textbf{\texttt{pedestal}}}, \textcolor{NavyBlue}{\textbf{\texttt{case}}}, \textcolor{NavyBlue}{\textbf{\texttt{step}}}, \\   \textcolor{red}{\textit{\texttt{windowpane}}}, \textcolor{NavyBlue}{\textbf{\texttt{ceiling}}}, \textcolor{red}{\textit{\texttt{sky}}}
\end{tabular}   \\ \hline
\begin{tabular}{l}\texttt{booth/}\\\texttt{indoor}\end{tabular} & \begin{tabular}{l}
\textcolor{NavyBlue}{\textbf{\texttt{poster}}}, \textcolor{NavyBlue}{\textbf{\texttt{pedestal}}}, \textcolor{NavyBlue}{\textbf{\texttt{partition}}}, \\   \textcolor{NavyBlue}{\textbf{\texttt{sales booth}}}, \textcolor{NavyBlue}{\textbf{\texttt{silver screen}}}, \\ \textcolor{NavyBlue}{\textbf{\texttt{jacket}}}
\end{tabular}    & \begin{tabular}{l}
 \textcolor{NavyBlue}{\textbf{\texttt{podium}}}, \textcolor{NavyBlue}{\textbf{\texttt{pedestal}}}, \textcolor{NavyBlue}{\textbf{\texttt{briefcase}}}, \\    \textcolor{NavyBlue}{\textbf{\texttt{spotlight}}}, \textcolor{red}{\textit{\texttt{windowpane}}}, \textcolor{NavyBlue}{\textbf{\texttt{person}}}
\end{tabular}  \\
\toprule
\end{tabular}}

\end{table}

\subsection{Generalization of the uninterpretable space}
\label{sec:beyond_single}
An advantage of our method is that the subspace $Uf$ contains information from $f$ not present in $A$. Thus, we can ask if the projection $U$ is generalizable, i.e, if it can be used to explain multiple models that are trained with the {same} feature representation. One can imagine such a scenario arising during transfer learning, when we use the feature space of a pre-trained model, and simply train a linear layer for the head of the model.  
Let $F^j = g^j o f$ for $j \in \{1, 2, \ldots m\}$. We want to produce a global explanation for each that uses the \textit{same} uninterpretable subspace $U$ of $f$, along with a subset of attributes $A$, specific for each head. 
In this case, we optimize
Eq.~\ref{eq:objection_single} over all the different $F^j$'s, keeping $U$ the same across all of these objectives, while $W^j_A$ and $V^j$ are learned separately for each model. 

We simulate a simple version of this by splitting the Places365 trained Resnet18 model into 16 models, each corresponding to one of the scene categories. Each of these models makes a fine-grained prediction on a new input. When examining predictions made by the  model on ADE20k, we find that for 2 scene categories (\texttt{outdoor-transportation} and \texttt{industrial-construction}), almost all images are predicted to be the same scene, and hence we drop these scene categories, using the remaining 14 as our different models. 

\begin{figure}[t]
\centering
\resizebox{\textwidth}{!}{
\begin{minipage}{.5\textwidth}
  \centering
  \includegraphics[width=\linewidth]{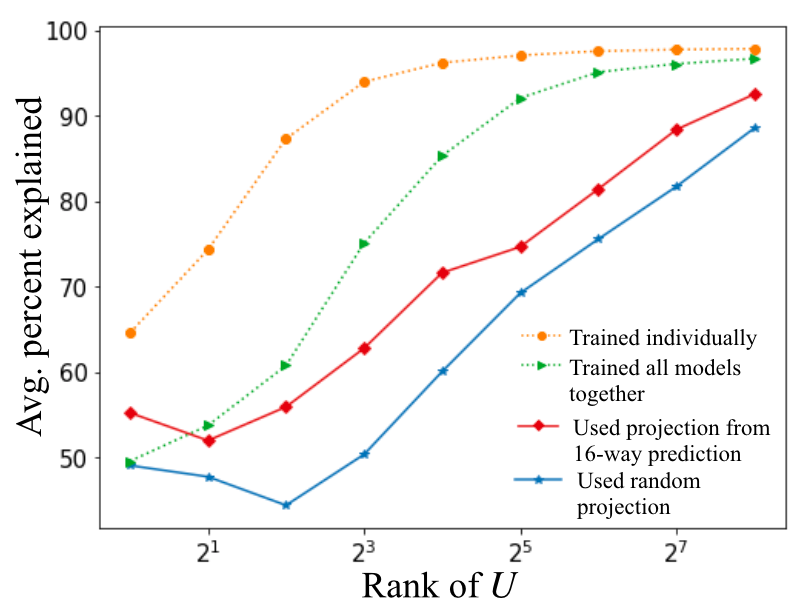}
\end{minipage}%
\begin{minipage}{.5\textwidth}
   \centering
   \includegraphics[width=\linewidth]{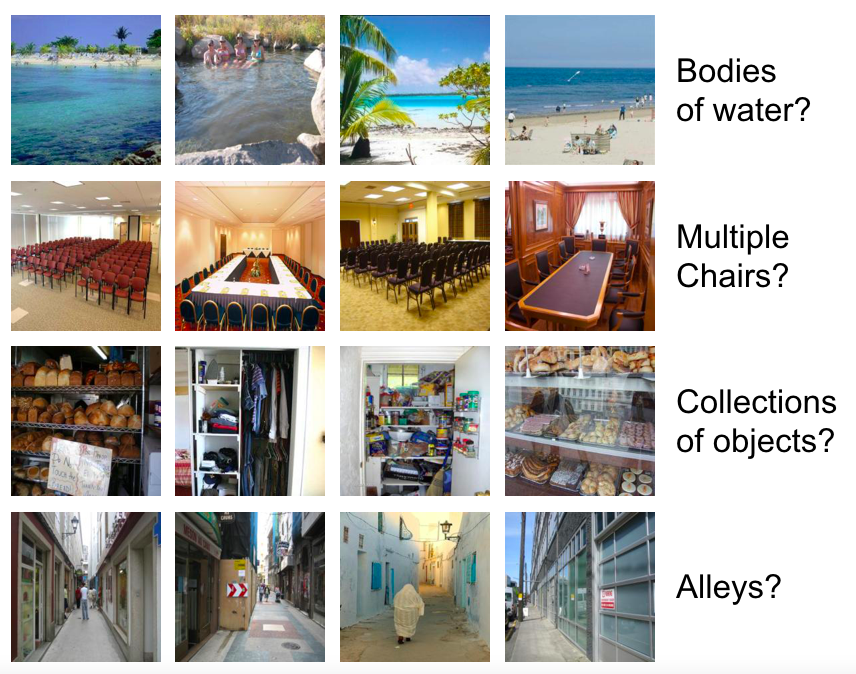}

\end{minipage}}
\caption{\emph{(left)} We compare the average fraction of the 14 fine-grained heads explained as we increase the rank when we learn the subspaces individually, all together, and using the low-rank subspace from the coarse explanation. We find that the learned subspace generalizes; the fraction explained is above that when using a random projection. (\emph{right}) The common subspace learned contains directions that appear to correspond to human-understandable concepts. Here, we visualize 4 such directions.} 
\label{fig:commonU}
\end{figure}

We first test the generalization of the subspaces learned for the 16-way scene classification model. We measure the percent of the model explained when using a subspace of rank $r$ learned for the 16-way scene classification and the set of attributes $A$. We compare this to learning a a separate low-rank subspace for each head, a common low-rank subspace across these 14 heads, as well as to a random projection $U_r$ of the feature space. Fig.~\ref{fig:commonU} (\emph{left}) shows this comparison. We see that while learning a subspace specifically for the fine grained heads does perform better, using the subspace from the coarse 16-way explanation performs well above the random projection, suggesting that the low rank subspace learned by ELUDE summarizes more of the model than a random projection of the feature space. 
For the common subspace learned using all 14 heads of the model, we visualize images that highly activate different axes, and find that (similar to Sec.~\ref{sec:places_16}), they appear to correspond to human understandable concepts. Fig.~\ref{fig:commonU} (\emph{right}) contains 4 such examples.  

\section{Explaining an interpretable-by-design model}
\label{sec:cub}
\noindent In this section, we apply ELUDE to explain a Concept Bottleneck model~\cite{koh2020conceptbottleneck}. This is an interpretable-by-design model by Koh*, Nguyen* and Tang* et. al, that focuses on predicting pre-labelled semantic attributes, and then using these attributes to make a final prediction, similar in spirit to our framework. An intermediate layer in the network has $K$ nodes, which are supervised to correspond to $K=112$ labelled attributes. The authors propose several methods for training such a model: (a) training the attribute predictor first and then the class predictor with the ground truth attribute labels, (b) training them separately, but using the outputs of the attribute predictor as inputs to the class predictor, and (c) training both jointly.
In our experiments, we use the trained weights provided by the authors for the \emph{joint} model with the concept loss weight = 1. This model is similar to our explanation, in so far as we predict a model's behavior by approximating it with a linear combination of the labelled semantic attributes (i.e., the trained attribute layer in this model will be the $f(x)$ layer we use in ELUDE).
However, rather than changing any model weights, we provide a complete post-hoc global explanation of the model, including additional unlabelled features. We explain this model because it allows us to understand how much information is actually captured by an interpretable-by-design model. 

\smallsec{Dataset} The Concept Bottleneck model is trained on the CUB dataset~\cite{WahCUB_200_2011}(license: unknown; non-commercial research/education use), which comprises of 200 different types of birds labelled with 312 attributes. This dataset has been  widely used in creating interpretable-by-design models, such as ProtoPNet~\cite{chen2018protopnet}, ProtoTree~\cite{nauta2021prototree} and Concept Bottleneck~\cite{koh2020conceptbottleneck}.

\smallsec{Implementation Details}
Since this model was trained on the CUB training set, we use the CUB test set with 5794 images to explain this model. We split these images randomly into training (60\%), validation (20\%) and test (20\%) subsets. Similar to Concept Bottleneck~\cite{koh2020conceptbottleneck}, we prune the set of 312 attributes, removing all those that occur in less than 10 images, giving us a subset of 112 attributes. The remaining details remain unchanged from Sec.~\ref{sec:places}. 

\smallsec{Explaining with labelled attributes} When training a linear classifier with the attributes $A$ to predict the Concept Bottleneck's output, only 29.4\% of the model's outputs are explained. The disparity between the low accuracy with the ground truth attributes $A(x)$ versus with the learned attribute layer $f(x)$ in Concept Bottleneck~\cite{koh2020conceptbottleneck} appears to be due to the artifact observed by the authors that the continuous values in the learned $f(x)$ are critical to the model's success (attempting to binarize the attribute predictions using a sigmoid layer led to a significant drop in accuracy). We find that the most common attributes considered important are related to the colours of different parts of the bird. A table with the important attributes is in the appendix. 

\begin{figure}[t]
    \centering
    \includegraphics[width=0.9\textwidth]{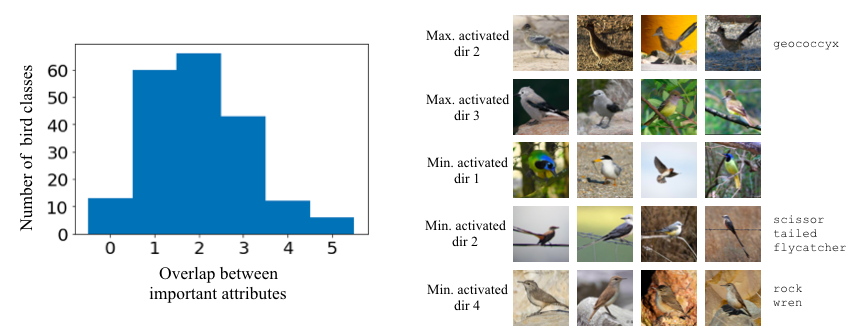}
    \caption{(\emph{left}) We view the histogram that corresponds to the overlap of important attributes between our learned explanation and the Concept Bottleneck model~\cite{koh2020conceptbottleneck}. We find that less than 3 attributes overlap for most classes. (\emph{right}) We show examples of images that minimally and maximally activate directions within the feature space. We find that certain directions in the rank 4 subspace correspond to specific birds, for example, images that maximally activate direction 2 correspond to \texttt{geococcyx}.}
    \label{fig:cub_cbn_U}
\end{figure}

Since the features of the Concept Bottleneck model refer to attributes from CUB, we compare the weights of the original model for these features with $W_A$. Surprisingly, we find that they do not correspond well. The 10 most important attributes for each bird class (based on the coefficient) for both our explanation as well as the original model, on average, overlap for 2 attributes. The full distribution of the overlap is shown in Fig.~\ref{fig:cub_cbn_U} (\emph{left}). One hypothesis for the difference in important attributes is that Concept Bottleneck models trade-off between learning attributes and the final class, leading to the neurons corresponding to these attributes encoding additional information.

\smallsec{Completing the explanation} 
Following ELUDE, we compute a low-rank projection $U$ from $f$ that allows us to explain the model's predictions to a greater degree. We find that with a rank of 8, we are able to explain over 80\% of the model's predictions. This suggests that while these 112 semantic attributes are not sufficient to explain the model, we only require 8 additional features to explain the model, and these are critical. We visualize the images that maximally and minimally activate the feature representations for ranks 1, 2, and 4 in Fig.~\ref{fig:cub_cbn_U}(\emph{right}). Interestingly, we start to see certain directions correspond to bird classes in rank 4, for example, images that maximally activate direction 2 correspond to \texttt{geococcyx}, images that minimally activate directions 2 and 4 correspond to \texttt{scissor-tailed-flycatcher} and \texttt{rock-wren} respectively. All these classes have a true positive rate of 0 in the attribute-only explanation, suggesting that the learned feature representation focuses on learning these classes. 

\smallsec{Explaining a standard Resnet18 model trained on CUB}
Finally, we evaluate whether Concept Bottleneck models rely more on attributes than a standard CNN, by explaining a standard Resnet18~\cite{he2016resnet} model trained on the CUB dataset. We find that we are able to explain 26.6\% of the model's prediction when using only the attributes, lower than 29.4\% of the Concept Bottleneck model. Moreover, when learning a low-rank feature representation from $f$, we find the percentage of the model explained is close to the upper bound only when $r=32$. The difference in the attribute-only accuracy as well as the rank between this and the Concept Bottleneck model suggests that this model relies less on the semantic attributes. Full results are in the {appendix}.


\section{Conclusions}
We propose ELUDE, a novel explanation framework, that allows us to predict the output of a black-box classification model based on a set of labelled attributes as well as a small set of unlabelled features. We demonstrate our method on two different settings and analyze the small set of unlabelled features to better understand what other visual cues the model might rely on. We stress that these other potential concepts (like the ``bowling-alleys'' or ``castle-like buildings'') are still very important --  one potential harm is future work misusing our framework to claim that their model and/or attribute set is sufficiently interpretable (e.g., because the unlabelled feature subspace $Uf(x)$ is low-dimensional). We also show that current interpretable-by-degin models are still far from being fully interpretable: when evaluating an interpretable-by-design Concept Bottleneck model, the labelled attributes explain less than 30\% of the entire model.   

\smallsec{Acknowledgements} This material is based upon work partially supported by the National Science Foundation under Grant No. 1763642. Any opinions, findings, and conclusions or recommendations expressed in this material are those of the author(s) and do not necessarily reflect the views of the National Science Foundation. We also acknowledge support from the Princeton SEAS Project X Fund, the Open Philanthropy grant and the Princeton SEAS Howard B. Wentz, Jr. Junior Faculty Award to OR. We thank the authors of \cite{zhou2018ibd,koh2020conceptbottleneck} for open-sourcing their code and models. We also thank Sharon Zhang for her initial work on this project and Moamen Elmassry, Jihoon Chung and the VisualAI lab for their feedback during the writing process. 

%
%
\bibliographystyle{splncs04}
\bibliography{visualai,references}


\appendix

In this supplementary material, we provide additional details about our experiments with both the Places365~\cite{zhou2017places} and CUB~\cite{WahCUB_200_2011} datasets

\section{Choice of $L1$ regularization}
\label{sec:l1_reg}
\setcounter{figure}{0}

\begin{figure}[t]
\centering
\resizebox{\textwidth}{!}{
\begin{minipage}{.5\textwidth}
  \centering
  \includegraphics[width=\linewidth]{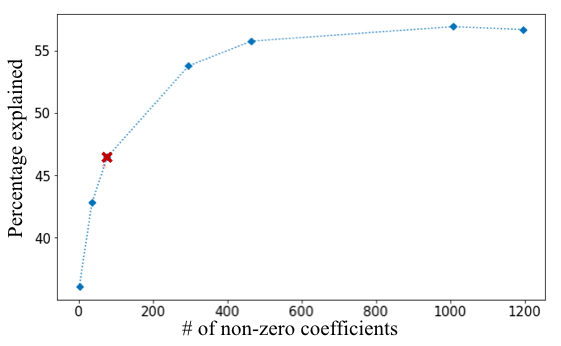}
\end{minipage}%
\begin{minipage}{.5\textwidth}
   \centering
   \includegraphics[width=\linewidth]{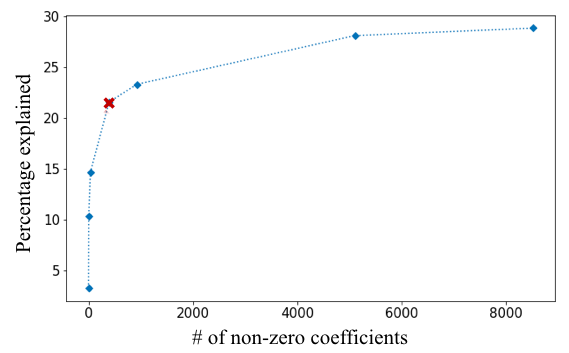}
\end{minipage}}
\begin{minipage}{.51\textwidth}
  \centering
  \includegraphics[width=\linewidth]{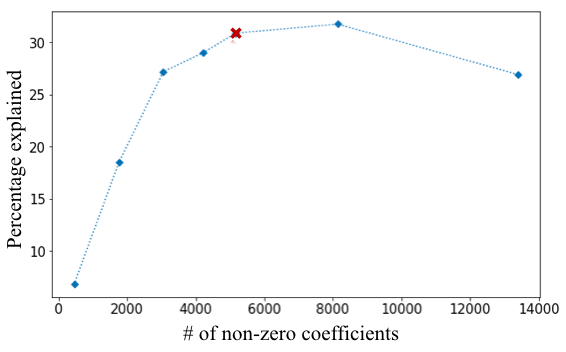}
\end{minipage}%
\begin{minipage}{.49\textwidth}
   \centering
   \includegraphics[width=\linewidth]{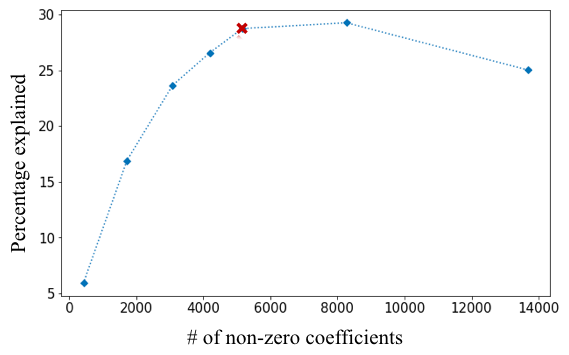}
\end{minipage}
\caption{We plot the tradeoff curves of the percentage of the model explained against the number of non-zero coefficients for all 4 models we explain: (\emph{top left}) Resnet18~\cite{he2016resnet} model on Places365~\cite{zhou2017places} predicting one of 16 classes, (\emph{top right}) Resnet18 model on Places365 predicting one of 365 classes, (\emph{bottom left}) Concept Bottleneck~\cite{koh2020conceptbottleneck} model on CUB~\cite{WahCUB_200_2011} and (\emph{bottom left}) Resnet18 model on CUB. Points marked in red were the hyperparameter chosen for $C$.} 
\label{fig:supp_places_lambda}
\end{figure}

When training the attribute-only explanations in sections 4 and 5 in the main text, we choose the regularization parameter based on the trade-off between the sparsity of the linear model and the percentage of the black box model explained. In Fig.~\ref{fig:supp_places_lambda}, we show this trade-off for all 4 models (Resnet18 model trained on Places365~\cite{zhou2017places} with 16 and 365 classes, Resnet18 model trained on CUB~\cite{WahCUB_200_2011} and the Concept Bottleneck model~\cite{koh2020conceptbottleneck} trained on CUB). In general, we picked the regularization hyperparameter based on the point at which the slope changed by a large extent (marked in red on the graphs). 
We chose the following values of $C$ (inverse regularization parameter for the Logistic Regression model from scikit-learn~\cite{scikit-learn}):

\begin{itemize}
    \item Resnet18 model trained on Places365 with 16 classes: 0.01
    \item  Resnet18 model trained on Places365 with 365 classes: 0.1
    \item Concept Bottleneck model trained on CUB: 0.4
    \item  Resnet18 model trained on CUB: 0.4
\end{itemize}

\section{Models trained on Places365: additional results}
\label{sec:supp_places}
\setcounter{figure}{0}
\setcounter{table}{0}

In this section, we present some additional results regarding our explanations for the Resnet18~\cite{he2016resnet} model trained on Places365~\cite{zhou2017places}. We show additional visualizations of the learned subspaces for both the 16-way and 365-way classification. We also present additional results showing that the learned subspace can generalize.

\smallsec{Explaining a binary classifier: additional results}
Here we list the actual coefficients of our explanation of the simple binary classifier that distinguishes between \texttt{indoor} and \texttt{outdoor} scenes. Table~\ref{tab:bin_weights} lists these coefficients, negative coefficients denote the attribute contributes towards \texttt{indoor} scenes, while positive coefficients denote that the attribute contributes towards \texttt{outdoor} scenes. 

\begin{table}[t]
    \centering
    \begin{tabular}{p{2cm}cc}
    \toprule
    Attribute  && Coefficient  \\
    \midrule
\texttt{sky} && $+ 2.14$ \\
\texttt{floor} && $- 1.59$ \\
\texttt{tree} && $+ 1.44$ \\
\texttt{wall} && $- 1.32$ \\
\texttt{building} && $+ 0.87$ \\
\texttt{ceiling} && $- 0.61$ \\
\bottomrule
    \end{tabular}
    \caption{We display the coefficients of the explanation learned for the binary scene classification model trained on Places365. Negative coefficients denote the attribute contributes towards \texttt{indoor} scenes, while positive coefficients denote that the attribute contributes towards \texttt{outdoor} scenes. }
    \label{tab:bin_weights}
\end{table}

\smallsec{Explaining a 16-way classifier: Additional visualizations}
In the Sec.~\ref{sec:places_16}, we saw that the GradCAM computed on the heatmap tended to focus on the edge of images, rather than the objects. Here, we verify that $(U^TV)^Tf(x)$ focuses on regions other than the salient objects. We compute the IOU of the heatmaps with the available object segmentation masks for ADE20k. To do so, we threshold each heatmap at the 75\% percentile, and compute the IOU of the thresholded heatmap with each object labelled in the image. For most objects (21/29), we find that the IOU is higher for the heatmap on the blackbox model. On average, the IOU for the blackbox heatmap is 8.8\%, compared to 6.5\% for the heatmap generated using our explanation. Moreover, the objects for which the heatmap for our explanation has a  higher IOU are generally background objects such as \texttt{wall}, \texttt{sky}, \texttt{floor}, \texttt{ceiling}, etc.

\begin{figure}[t]
    \centering
    \includegraphics[width=\textwidth]{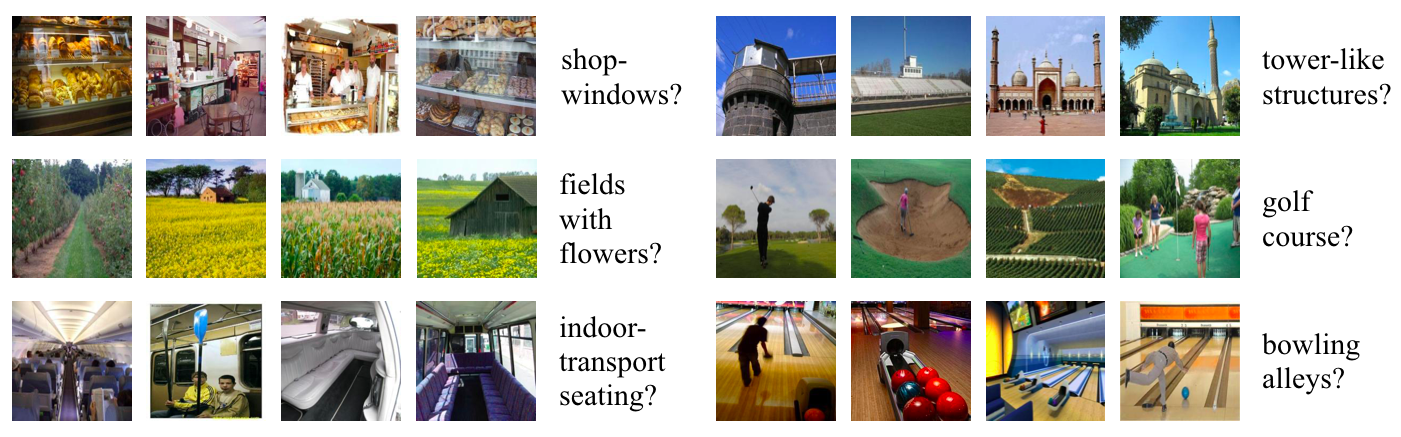}
    \caption{We visualize some directions within the rank 8 subspace that we learn to explain a 16-way Places365 trained classifier. We see that these correspond to human understandable concepts, like shop-windows or golf courses.}
    \label{fig:rank8}
\end{figure}

\begin{figure}[t]
    \centering
    \includegraphics[width=\textwidth]{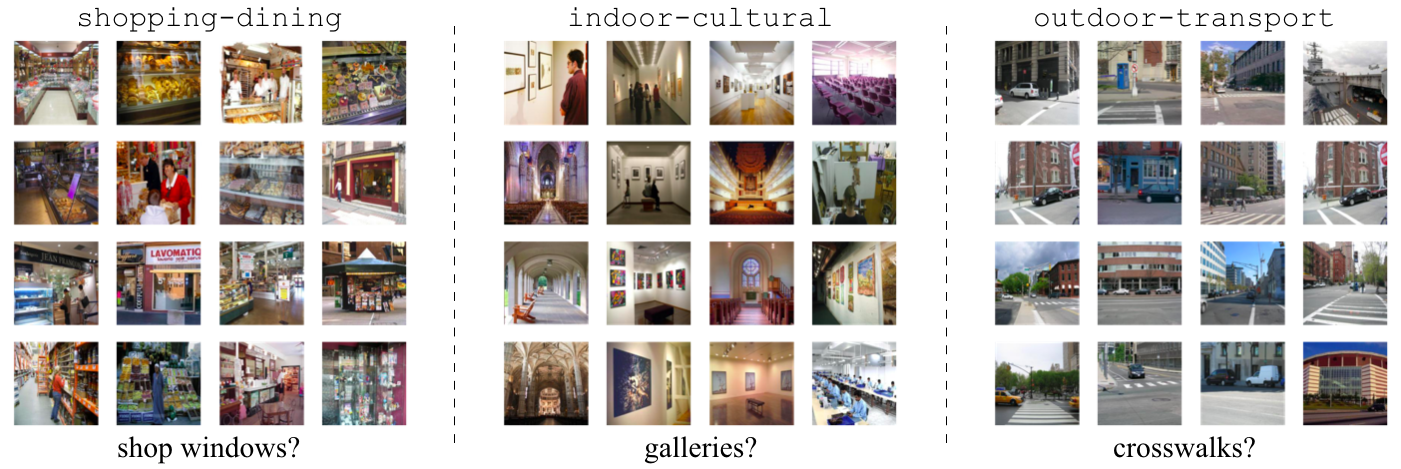}
    \caption{We visualize images that have the highest scores of $U^TVf(x)$ for different classes. For some classes we can identify additional concepts that might be useful for the class, for example, `shop-windows' for \texttt{shopping-dining}. However, for others, we see that this just picks up subclasses within the class, for example, something like `gallery' for \texttt{indoor-cultural}. }
    \label{fig:per_class_act}
\end{figure}

In the main paper, we visualized images that highly activate directions within the rank 4 subspace. Here, we additionally visualize some directions within the rank 8 subspace (Fig.~\ref{fig:rank8}). Another method we use to understand this low-rank subspace is to consider images that have the largest score of $U^TVf(x)$ for different classes (Fig.~\ref{fig:per_class_act}). For some classes we can identify additional concepts that might be useful for the class, for example, `shop-windows' for \texttt{shopping-dining}. However, for others, we see that this just picks up subclasses within the class, for example, something like `gallery' for \texttt{indoor-cultural}.

\smallsec{Explaining a 365-way classifier: Additional results} We listed some of the important attributes found by ELUDE compared to Interpretable Basis Decomposition~\cite{zhou2018ibd} in Table~2 in the main text. Here, we show the change in the percentage of the model explained as well as visualize the directions learned for the lower rank subspaces.  

\begin{figure}[t!]
\centering
\begin{minipage}{.5\textwidth}
  \centering
  \begin{tabular}{>{\centering\arraybackslash}p{2cm} c p{0.25cm} >{\centering\arraybackslash}p{2cm} c}
\toprule
attr. name & count && attr. name & count \\
\toprule
\texttt{sky} & 81 && \texttt{table} & 34 \\
\texttt{floor} & 62 && \texttt{ceiling} & 30 \\
\texttt{building} & 53 && \texttt{painting} & 30\\
\texttt{tree} & 52 && \texttt{plant} & 29 \\
\texttt{person} & 48 && \texttt{car} & 23 \\
\texttt{road} & 43 && \texttt{grass} & 23\\
\texttt{wall} & 39 && \texttt{windowpane} & 22  \\
\toprule
\end{tabular}

\end{minipage}%
\begin{minipage}{.5\textwidth}
   \centering
\includegraphics[width=\linewidth]{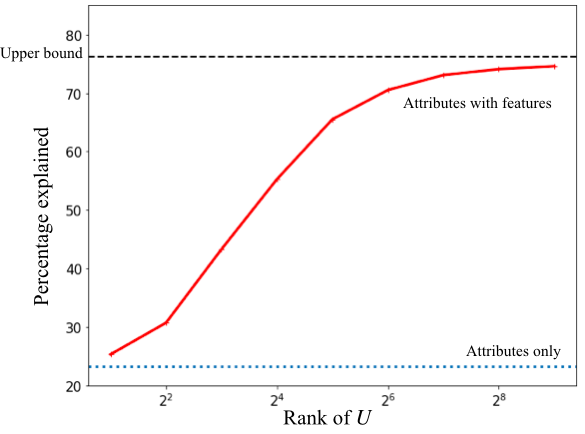}
\end{minipage}
\caption{\emph{(left)} This table shows the number of times an attribute is among the top 10 most important attributes for each of the 365 classes in Places365. The important attributes seem to similar to those when predicting 16 classes. (\emph{right}) We view the trade-off between the rank of the feature representation and the percentage of the Resnet18 model that is explained by this representation. We find that at a rank of 32, the subspace is able to explain over 65\% of the model, and at a rank of 64, this number increases to over 70\%.}
\label{suppfig:365_analysis}
\end{figure}

To start, we consider the true positive rates among the various classes. Out of the 121 scenes that are predicted more than 25 times within ADE20k~\cite{zhou2017ade20k}, 46 of these scenes are not predicted using our explanation on just the attributes, suggesting that we still have a way to go before explaining the model with just the attributes. Some of these classes include \texttt{hotel-outdoor}, \texttt{berth} and \texttt{arcade}. Among the 8 scenes that are well explained using these attributes (true positive rate $>$ 0.8), we have \texttt{homeoffice}, \texttt{bedroom}, \texttt{kitchen}, \texttt{diningroom}, \texttt{bathroom}, \texttt{highway}, \texttt{street} and \texttt{skyscraper}. We find the attributes that are featured among the top 10 attributes for several classes. These include attributes like \texttt{sky}, \texttt{floor}, \texttt{building}, \texttt{tree} and  \texttt{person}. Surprisingly, we find that the important attributes (such as \texttt{sky}, \texttt{wall}, \texttt{ceiling}) remain the same across explaining this model at different levels of 
granularity. Fig~\ref{suppfig:365_analysis} (\emph{left}) lists the most frequent attributes.

We also view the fraction of this model explained as we change the rank of $U$ (Fig.~\ref{suppfig:365_analysis}(\emph{right})), and find that we are able to explain over 65\% of the model at a rank of 32, and this number increases to over 70\% at a rank of 64. We see that the complexity of this model manifests both in the rank of the subspace required to well explain the model, as well as the fraction of the model that is explained by the attributes alone.

\begin{figure}[t]
    \centering
    \includegraphics[width=\linewidth]{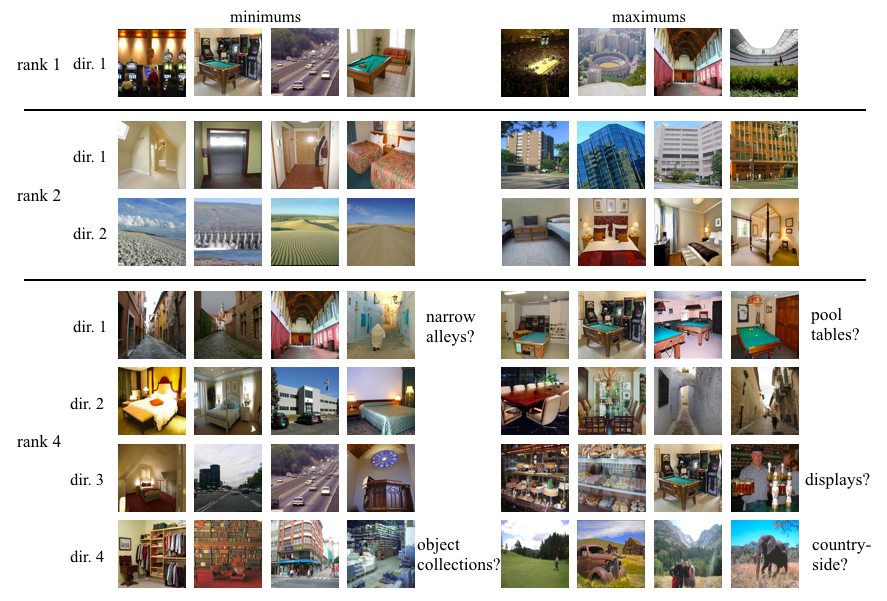}
    \caption{We show examples of images that highly activate directions within the low-rank subspace. As the rank increases, we see that some directions appear to correspond to human understandable concepts. For example, with the rank 4 projection, images that minimally activate direction 4 correspond to collections of objects.}
    \label{suppfig:365_activations}
\end{figure}

We also view the directions that correspond to the different directions learned for the lower rank subspaces learned, as shown in Fig~\ref{suppfig:365_activations}. We note that some directions within the rank 4 subspace appear to correspond to certain concepts, for example, the images that minimally activate direction 4 appear to be collections of objects. Since the rank 4 space is still insufficient to well-explain this model, we see the there are several directions that still appear to correspond to multiple concepts. 

\begin{figure}[t]
\centering
\includegraphics[width=0.62\textwidth]{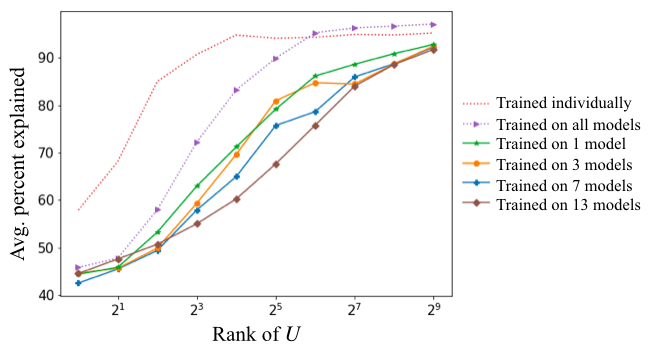}
\caption{We understand how well this learned subspace $U$ can generalize to unseen models, by training and testing on different models. As an upperbound we train a feature representation using all 14 models. With a slightly larger rank, we are able to explain a similar percentage of the model when training with at least 3 other models. }
\label{fig:unknown_model}
\end{figure}

\smallsec{Generalization: additional results}
We also test how well the learned subspace for the 14 fine-grained heads generalizes to the other heads. We train different feature representations using different numbers of these models, and report the percentage of the model explained with the same low-rank subspace $Uf$ for the unseen models. Fig.~\ref{fig:unknown_model} shows the average percentage of all model explained for all the models that are unseen at training time. We note that the subspace $U$ does generalize across models : a rank 64 $U$ trained on just 3 models is able to explain all models as well as a rank 16 $U$ trained on all models

\section{Models trained on CUB: additional results}
\setcounter{figure}{0}
\setcounter{table}{0}
\label{sec:supp_cub}

In this section we present some of the additional results when we run ELUDE on the two models trained on the CUB~\cite{WahCUB_200_2011} dataset: the Concept Bottleneck model~\cite{koh2020conceptbottleneck} and a baseline Resnet18~\cite{he2016resnet} model. 

\begin{figure}[ht!]
\centering

\begin{minipage}{.5\textwidth}
 \centering
\includegraphics[width=\linewidth]{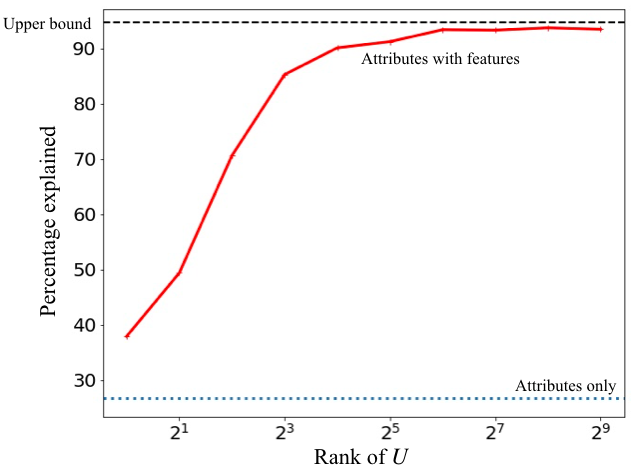}
\end{minipage}%
\begin{minipage}{.5\textwidth}
   \centering
\includegraphics[width=\linewidth]{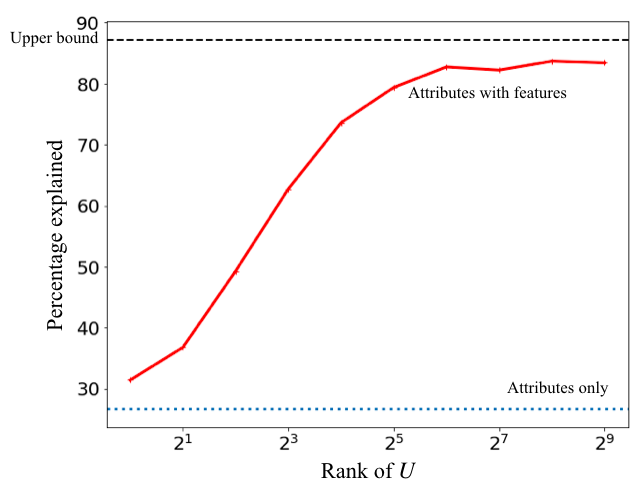}
\end{minipage}
\caption{We view how the fraction of the model explained varies by the rank of $U$ for  curves of the fraction of the model explained  the concept bottleneck model (\emph{left}) and a baseline Resnet18 model (\emph{right}) trained on CUB. TO explain over 80\% of the model, the Concept Bottleneck model requires only 8 additional features whereas the baseline model requires 32.}
\label{suppfig:cub_tradeoff}
\end{figure}

\begin{figure}[ht]
    \centering
    \includegraphics[width=\textwidth]{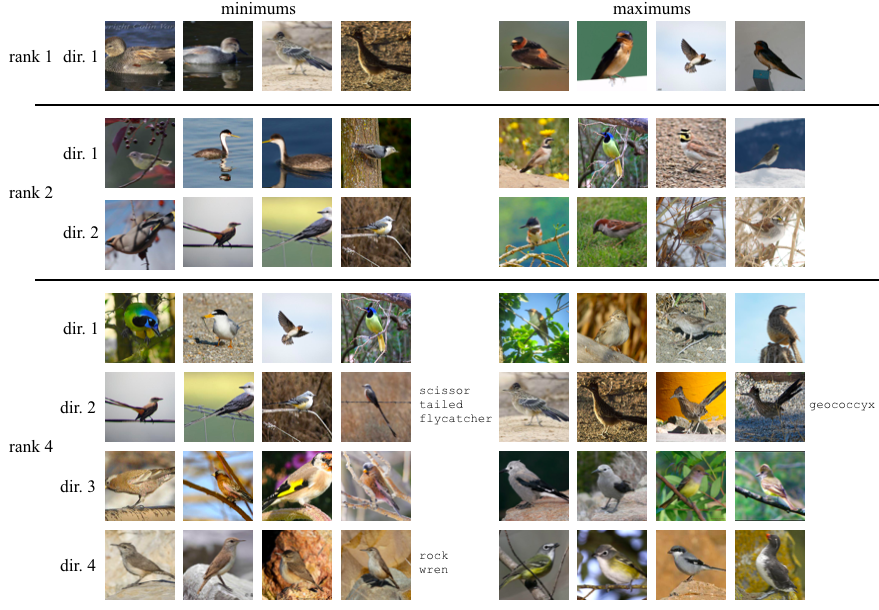}
\\
    \hrule 
    \includegraphics[width=\textwidth]{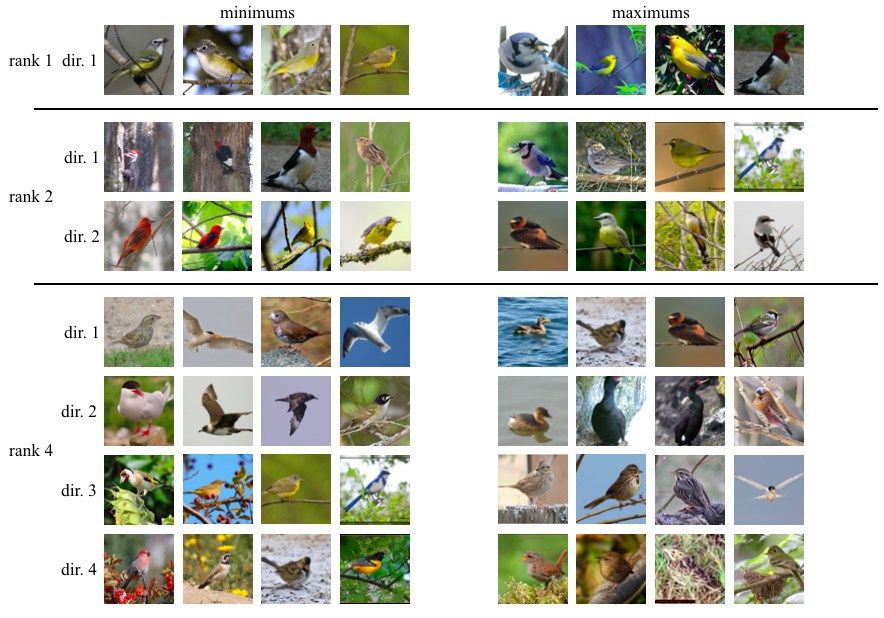}
    \caption{Images that maximally (\emph{right}) and minimally (\emph{left}) activate directions for subspaces of different ranks computed for the Concept Bottleneck model and for the baseline Resnet18 model.}
    \label{suppfig:cub_activations}
\end{figure}

We first view the number of times an attribute is among the top 10 most important attributes for each of the 200 bird classes in CUB for the two models in Tab~\ref{supptab:cub_attr}. We also show the trade-off curves between the ranks of $U$ used and the fraction of the model explained in Fig~\ref{suppfig:cub_tradeoff}. Finally, we show the images that maximally and minimally activate directions for subspaces of ranks 1, 2 and 4 for both explanations learned. We see that certain directions start to correspond to bird classes for the Concept Bottleneck model for rank 4, whereas the directions still correspond to multiple concepts for the baseline Resnet18 model (Fig.~\ref{suppfig:cub_activations}).

\begin{table}[ht!]
\centering
\caption{These tables shows the number of times an attribute is among the top 10 most important attributes for each of the 200 bird species for the concept bottleneck model (\emph{left}) and a baseline Resnet18 model (\emph{right}) trained on CUB. The important attributes seem to be colours of various body parts of the bird, along with the bill-shape and size.}
\label{supptab:cub_attr}
\begin{minipage}{.5\textwidth}
 
\begin{tabular}{c c}
\toprule
attribute name & count \\
\toprule
\texttt{eye-color::buff} & 69\\
\texttt{bill-color::orange} & 56\\
\texttt{shape::swallow-like} & 53\\
\texttt{bill-length::longer-than-head} & 53\\
\texttt{throat-color::black} & 51\\
\texttt{crown-color::orange} & 51\\
\texttt{underparts-color::black} & 49\\
\texttt{bill-shape::spatulate} & 45 \\
\texttt{head-pattern::eyering} & 42\\
\texttt{tail-pattern::multi-colored} & 40 \\
\toprule
\end{tabular}

\end{minipage}%
\begin{minipage}{.5\textwidth}
   \centering
 \begin{tabular}{c c}
\toprule
attribute name & count \\
\toprule
\texttt{eye-color::buff} & 62 \\
\texttt{bill-length::longer-than-head} & 56 \\
\texttt{crown-color::orange} & 54 \\
\texttt{bill-color::orange} & 51 \\
\texttt{shape::swallow-like} & 50 \\
\texttt{throat-color::black} & 50 \\
\texttt{underparts-color::buff} & 48 \\
\texttt{head-pattern::eyering} & 47 \\
\texttt{underparts-color::black} & 46 \\
\texttt{bill-shape::spatulate} & 41 \\
\toprule
\end{tabular}

\end{minipage}
\end{table}

\end{document}